\documentclass[runningheads]{llncs}
\usepackage[T1]{fontenc}
\usepackage{graphicx} 
\usepackage{amsmath}
\usepackage{array}
\usepackage{adjustbox}
\usepackage{multirow}
\usepackage{booktabs} 

\usepackage[colorlinks,linkcolor=blue,citecolor=red]{hyperref}
\usepackage{wrapfig}
\usepackage{makecell}
\usepackage{float}
\usepackage{graphicx}
\usepackage{caption} 
\title{EU-Nets: Enhanced, Explainable and Parsimonious U-Nets}

\author{Bohang Sun \and Pietro Liò}
\authorrunning{B. Sun and P. Liò}
\institute{
School of Information and Software Engineering, University of Electronic Science and Technology of China \
\email{bobsun@std.uestc.edu.cn}
\and
Department of Computer Science and Technology, University of Cambridge \
\email{pl219@cam.ac.uk}
}

\begin{document}

\maketitle

\begin{abstract}
In this study, we propose MHEX+, a framework adaptable to any U-Net architecture. Built upon MHEX+, we introduce novel U-Net variants, EU-Nets, which enhance explainability and uncertainty estimation, addressing the limitations of traditional U-Net models while improving performance and stability. A key innovation is the Equivalent Convolutional Kernel, which unifies consecutive convolutional layers, boosting interpretability. For uncertainty estimation, we propose the collaboration gradient approach, measuring gradient consistency across decoder layers. Notably, EU-Nets achieve an average accuracy improvement of 1.389\% and a variance reduction of 0.83\% across all networks and datasets in our experiments, requiring fewer than 0.1M parameters.
\end{abstract}
\section{Introduction}

In clinical practice, accurate segmentation of anatomical structures or lesions is crucial for diagnosis, treatment planning, and prognosis. However, deep neural networks, despite their strong performance, often lack transparency, making their decision-making process difficult to understand.\cite{marinov2024deepinteractivesegmentationmedical,XAIinmedical} 

Existing explainability methods, particularly Grad-CAM and its variants~\cite{wang2020score,chattopadhay2018grad,selvaraju2017grad}, have demonstrated effectiveness in classification tasks. However, their application to semantic segmentation is limited, as they primarily highlight the predicted regions without explaining the underlying decision-making process. \cite{gradisnotexplainable,gradisnotexplainable2,dodont}.

Beyond explainability, uncertainty estimation is also crucial in medical image segmentation. Inference methods based on uncertainty modeling, such as Subjective Logic (SL)~\cite{subjectiveLogic} and Dempster-Shafer Theory (DST)~\cite{DempsterShafer}, can quantify belief mass. However, these methods typically require training models from scratch, which may degrade accuracy, especially in cases where the loss function is sensitive. Monte Carlo Dropout (MC Dropout)~\cite{gal2016dropout} requires the model to incorporate dropout layers at inference, while Deep Ensembles~\cite{lakshminarayanan2017simplescalablepredictiveuncertainty} estimate uncertainty by measuring model disagreement but incur high computational costs due to the need for training multiple identical or diverse models.  

Motivated by the above considerations, we propose a novel explainable U-Net variant, referred to as \emph{EU-Nets}, based on the Multi-Head Explainer (MHEX) framework~\cite{mhex}.  The main contributions of this work are summarized as follows:

\begin{itemize}
     \item \textbf{Enhanced U-Net Variants.} We propose the MHEX\texttt{+} framework, which is applicable to any U-Net architecture, achieving higher accuracy and stability with minimal parameter overhead (<0.1M).
    \item \textbf{Revealing the Decision-Making Process.} We introduce an equivalent convolutional kernel that merges consecutive convolutions into a single operation, improving interpretability.
    \item \textbf{Reliable Uncertainty Estimation.} We propose the collaboration gradient method, which provides a trustworthy uncertainty measure by examining whether backpropagated gradients reach consensus at given pixel.
\end{itemize}

\section{Related Work}

\paragraph{U-Net and Its Variants}  
U-Net~\cite{unet} introduced skip connections to bridge low-level and high-level features, significantly improving segmentation performance. U-Net++~\cite{unet++} further enhanced this design by incorporating dense connections and deep supervision to refine feature aggregation. AHF-U-Net~\cite{AHF-U-Net} employed attention mechanisms similar to SE~\cite{hu2018squeeze} and CBAM~\cite{woo2018cbam} to compute importance coefficients for fusion connections, enhancing feature selection adaptively. U-Net Transformer~\cite{U-NetTransformer} integrated self-attention in the bottleneck and cross-attention in skip connections.

\paragraph{Explainability Methods}  
Grad-CAM and its variants~\cite{wang2020score,chattopadhay2018grad,selvaraju2017grad} generate saliency maps by weighting feature maps based on gradients, second-order derivatives, or model confidence. SHAP~\cite{lundberg2017unified} estimates feature importance by perturbing inputs and computing Shapley values. LIME~\cite{ribeiro2016whyitrustyou} approximates local decision boundaries by training interpretable linear models around the input space. MHEX (Multi-Head Explainer)~\cite{mhex} enhances both performance and explainability through lightweight modules and fine-tuning while also providing uncertainty estimation.  

\paragraph{Uncertainty Estimation}  
Subjective Logic (SL)~\cite{subjectiveLogic} and Dempster-Shafer Theory (DST)~\cite{DempsterShafer} encourage models to quantify uncertainty in predictions. Monte Carlo Dropout (MC Dropout)~\cite{gal2016dropout} estimates uncertainty by enabling dropout at inference time. Deep Ensembles~\cite{lakshminarayanan2017simplescalablepredictiveuncertainty} aggregate predictions from multiple independently trained models to improve robustness. Test-Time Augmentation (TTA)~\cite{TTA} leverages augmented inputs during inference to indirectly reflect model uncertainty. MHEX\cite{mhex} introduces an alternative uncertainty estimation approach by analyzing gradient consistency. 

\section{Method}
\subsection{EU-Nets}

MHEX has been shown to enhance interpretability and uncertainty estimation in classification tasks. However, its integration into semantic segmentation remains underexplored. 

The original MHEX framework \cite{mhex} employs deep supervision~\cite{lee2014deeplysupervisednets,li2022comprehensivereviewdeepsupervision} and attention gating mechanisms~\cite{schlemper2019attention} to enhance performance. During the explanation phase, an equivalent transformation matrix \( W_{\text{equiv}} = W_1 W_2 \) is constructed.

In this work, we introduce MHEX+, an extended version of MHEX tailored for semantic segmentation. MHEX+ is designed to be applicable to any U-Net architecture. To validate its broad effectiveness, we integrate it into U-Net, U-Net++, U-Net Transformer, and AHF-U-Net, forming a new family of explainable architectures: EU-Nets.

\subsection{MHEX+ Module}
\label{sec:MHEX_Module}

\begin{figure}
    \centering
    \includegraphics[width=0.8\linewidth]{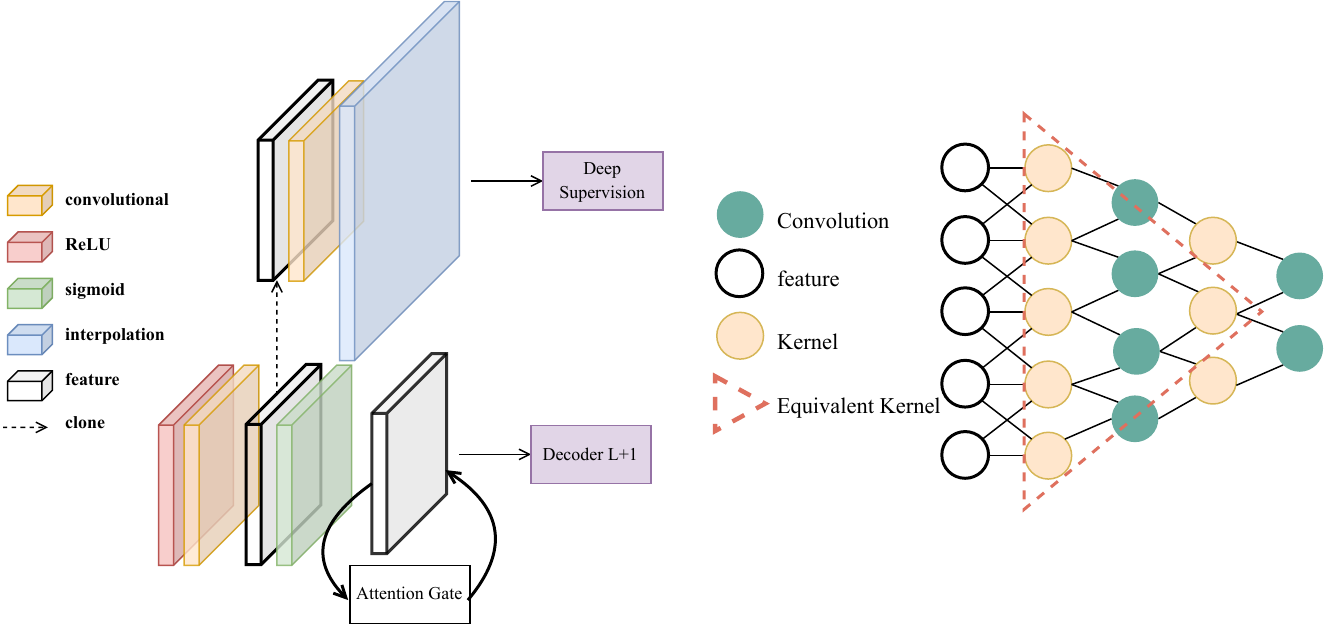}
    \caption{The left side illustrates the working mechanism of the MHEX+ module, while the right side represents the Equivalent Convolutional Kernel}
    \label{fig:mhex_equivConv}
\end{figure}

In EU-Nets, the MHEX+ module operates as follows:
\[
   Y =  \text{ReLU}(\text{Conv}_1(X) + b_1)
\]
\[
   \text{Attention Gate} = \sigma(Y), \quad Y_{\text{attended}} = \sigma(Y) \odot Y
\]
\[
   \text{Deep Prediction} = \text{Conv}_2(Y) + b_2
\]
where \( \text{Conv}_1 \) is a \( 1 \times 1 \) convolution followed by ReLU activation, and \( \sigma(\cdot) \) denotes the sigmoid function applied in the attention gate. The final deep prediction is computed via \( \text{Conv}_2 \). For simplicity, biases are omitted in our implementation.

The MHEX+ module in EU-Nets is depicted on the left side of Figure~\ref{fig:mhex_equivConv}.

\subsection{Equivalent Convolutional Kernel \& Salience Map}
\label{sec:Equivalent Convolutional Kernel}

We extend the \(W_{equiv}\) principle to convolutions. Given two consecutive layers \( \text{Conv}_1 \) and \( \text{Conv}_2 \), their equivalent kernel is computed as:
\[
    W_{\text{equiv}}[c] = \sum_{j=1}^{C_{\text{in}}} W_{\text{Conv}_2}[c, j] \cdot W_{\text{Conv}_1}[j]
\]
where \( c \) represents the output channel corresponding to a specific class. \( \text{Conv}_1 \) is part of the Attention Gate, while \( \text{Conv}_2 \) is responsible for deep prediction.

In this way, multiple convolutional operations are consolidated into a single equivalent transformation. The Equivalent Convolutional Kernel enables efficient Class Activation Map (CAM) generation:
\[
    \text{CAM}_c = \sum_{j=1}^{C_{\text{in}}} W_{\text{equiv}}[c, j] \cdot A[j]
\]
where \( A[j] \) represents the activation map of the \( j \)-th channel.

It is important to note that the time complexity of Grad-CAM is \(O(n^2 \cdot C)\), whereas MHEX+CAM has a time complexity of \(O(C)\), where \(C\) denotes the \(num_\text{channels}\), which is a constant usually.

The right side of Figure~\ref{fig:mhex_equivConv} illustrates the Equivalent Convolutional Kernel.

\subsection{Integration}
\label{sec:Integration}
U-Net-like networks consist of three stages: downsampling (encoder), extracting low-level features; bottleneck, capturing global context; and upsampling (decoder), reconstructing the segmentation map.

\begin{figure}
    \centering
    \includegraphics[width=1\linewidth]{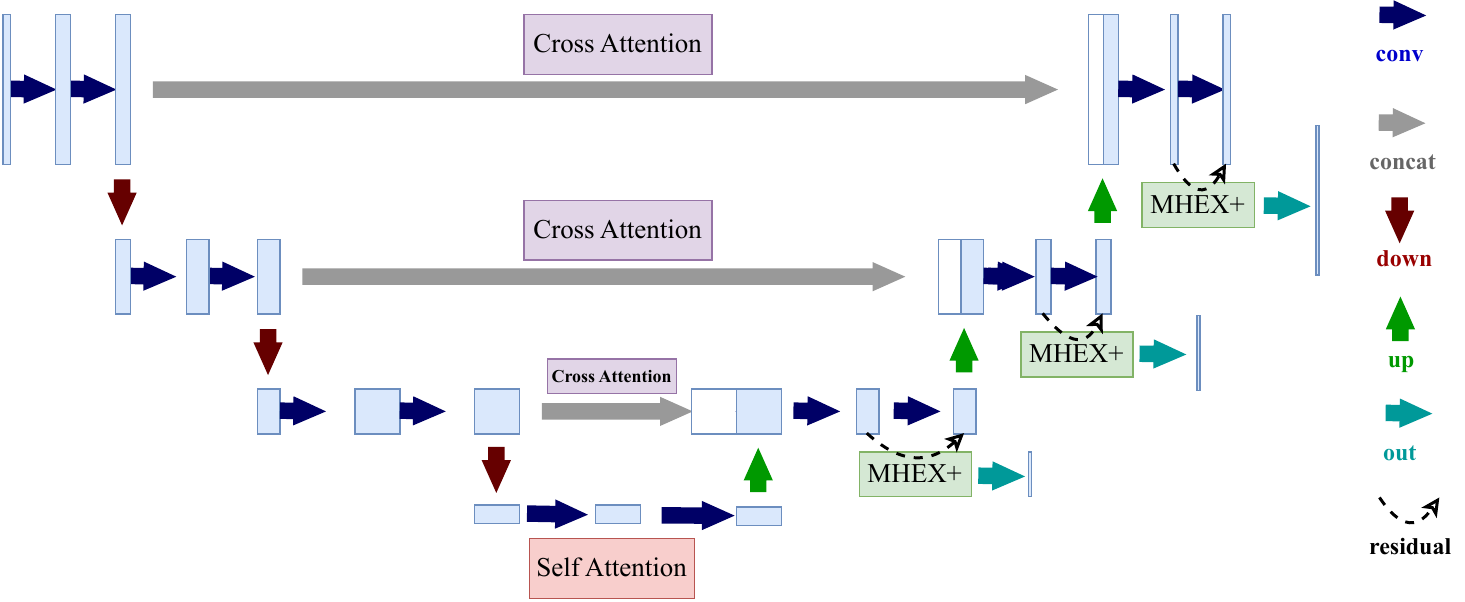}
    \caption{Example of MHEX+ integration with U-Net Transformer (EU-Net Transformer)}
    \label{fig:EUNT}
\end{figure}

We construct residual connections in the double convolution module across all decoders in U-Nets to integrate MHEX+, following the structural design principles of ResNet~\cite{he2015deepresiduallearningimage}. For U-Net++, we use the densely connected version rather than the deep supervision version, as MHEX+ already incorporates deep supervision. Please refer to Figure~\ref{fig:EUNT}.

\subsection{Uncertainty Analysis}
\label{sec:CollabGrad}

In this section, we introduce the concept of \textbf{collaboration gradient} to measure the consistency between adjacent MHEX+ blocks at the pixel level. 

We compute gradients for each pixel \((i, j)\) and measure the cosine similarity of gradient vectors from adjacent MHEX+ blocks at layers \(l\) and \(l+1\) in EU-Nets decoders. Specifically, for each pixel, the Uncertainty (\(U^{(i,j)}\)) is defined as:

\[
U^{(i,j)} = \sum_{L-1} \cos(\nabla_{l}^{(i,j)}, \nabla_{l+1}^{(i,j)}) 
= \frac{\nabla_{l}^{(i,j)} \cdot \nabla_{l+1}^{(i,j)}}{\|\nabla_{l}^{(i,j)}\| \|\nabla_{l+1}^{(i,j)}\| + \epsilon},
\]

\begin{itemize}
    \item \(\nabla_{l}^{(i,j)} = \frac{\partial \mathcal{L}_{l}^{(i,j)}}{\partial C_1}\) denotes the gradient of the cross-entropy loss \(\mathcal{L}_{l}^{(i,j)}\) with respect to the parameter \(C_1\) at layer \(l\);
    \item \(\nabla_{l+1}^{(i,j)} = \frac{\partial \mathcal{L}_{l+1}^{(i,j)}}{\partial C_1}\) represents the corresponding gradient from layer \(l+1\);
    \item \(\cos(\nabla_{l}^{(i,j)}, \nabla_{l+1}^{(i,j)})\) represents the cosine similarity between the two gradient vectors at pixel \((i,j)\).
\end{itemize}

To verify the validity of our proposed uncertainty measure, we compare it against the \textbf{Deep Ensemble} method \cite{lakshminarayanan2017simplescalablepredictiveuncertainty}, which constructs multiple independent models \(\{M_1, M_2, ..., M_N\}\) and estimates uncertainty for a given pixel \((i, j)\) based on the variation in predicted probabilities across models. Two commonly used uncertainty metrics in Deep Ensemble are:
\[
\sigma^2(y = c | x) = \frac{1}{N} \sum_{m=1}^{N} (P_m(y = c | x) - P(y = c | x))^2
\]
\[
H(y | x) = -\sum_{c} P(y = c | x) \log P(y = c | x)
\]
where \(P_m(y = c | x)\) is the probability assigned by model \(M_m\), and \(P(y = c | x)\) is the mean probability across models. Higher entropy \(H(y | x)\) indicates greater uncertainty due to model disagreement.

\section{Experiments}

\subsection{Datasets}
\label{sec:Datasets}
To evaluate the performance of EU-Nets, we tested them on four diverse datasets spanning multiple imaging modalities (MRI, CT, Ultrasound, and Dermoscopic images). Each dataset presents unique segmentation challenges:

\begin{itemize}
    \item \textbf{MSD-Heart} \cite{antonelli2022medical}: 30 MRI scans (320 × 320), left atrium segmentation, limited sample size, high anatomical variability.
    \item \textbf{MSD-Spleen} \cite{antonelli2022medical}: 61 CT scans (500 × 500), spleen segmentation, varied field-of-view.
    \item \textbf{Breast Ultrasound} \cite{al2020dataset}: 780 ultrasound images (600 patients, 500 × 500), classification: normal, benign, malignant, multi-label cases merged manually.
    \item \textbf{ISIC 2016} \cite{codella2018skin}: ~900 training, 350 test dermoscopic images (1024 × 768), melanoma segmentation with expert-annotated masks.
\end{itemize}

\begin{figure}[p]
\centering
\begin{minipage}{\textwidth}
\captionsetup{type=table}
    \centering
    \begin{tabular}{l c l l l l}
    \toprule
    Model                  & \#Params & MSD-Heart$^D$            & MSD-Spleen$^C$           & ISIC-Skin$^C$        & Breast $^D$              \\ 
    \midrule
    UN                  & 7.76M           & 86.72 ± 2.62        & 89.19 ± 3.12        & 84.62 ± 0.65         & 68.71 ± 4.24        \\ 
    EUN             & 7.85M           & 87.24 ± 2.37         & 91.18 ± 2.14        & 85.83 ± 0.40         & 68.19 ± 3.99        \\ \midrule
    UN++                & 9.16M           & 86.04 ± 1.76         & 89.38 ± 4.31        & 84.71 ± 0.16         & 70.10 ± 3.56        \\ 
    EUN++           & 9.25M           & 87.94 ± 2.29         & 91.32 ± 2.04        & 85.83 ± 0.10         & 71.99 ± 2.53        \\ \midrule
    AHF-UN              & 9.86M          & 76.12 ± 7.33$^*$        & 89.89 ± 5.19        & 84.03 ± 1.30         & 75.77 ± 3.23        \\ 
    AHF-EUN         & 9.95M           & 86.41 ± 2.72         & 89.67 ± 4.58        & 85.30 ± 0.96         & 73.05 ± 3.10        \\ \midrule
    UN-T            & 9.42M           & $82.92 \pm 2.90$         & 89.56 ± 3.51        & 81.36 ± 1.80         & 76.38 ± 3.92        \\ 
    EUN-T       & 9.51M           & 87.10 ± 2.25         & 85.60 ± 2.97        & 79.33 ± 0.38         & 77.96 ± 2.48        \\ 
    \bottomrule
    \end{tabular}
    \caption{Comparison of segmentation performance using Dice coefficient (\%). The most suitable loss function was chosen for each dataset: $^D$ denotes the use of Dice loss, $^C$ indicates cross-entropy loss, and $^*$ signifies cases where cross-entropy loss was used due to training failure.}
    \label{tab:comparison}
\end{minipage}

\vspace{1cm} 

\begin{minipage}{\textwidth}
\centering
\setlength{\tabcolsep}{1pt} 
\begin{tabular}{ccccccccc}
 & \multicolumn{4}{c}{\includegraphics[width=0.28\textwidth]{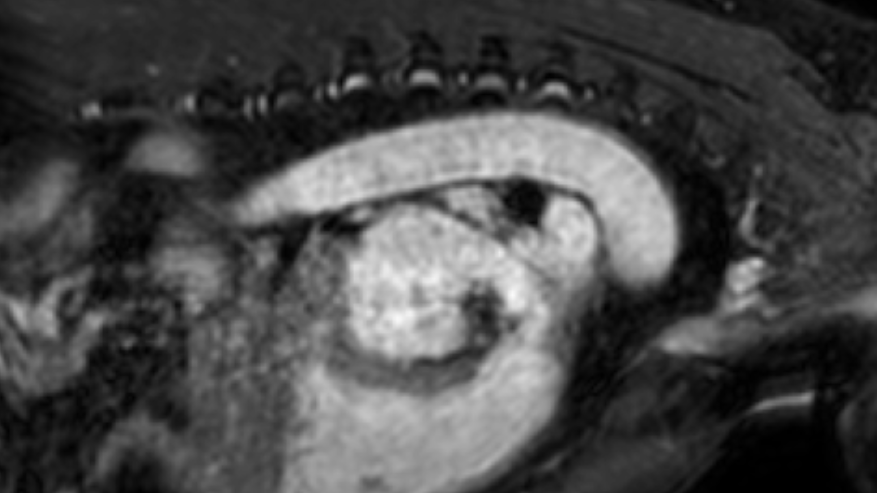}} & \multicolumn{4}{c}{\includegraphics[width=0.28\textwidth]{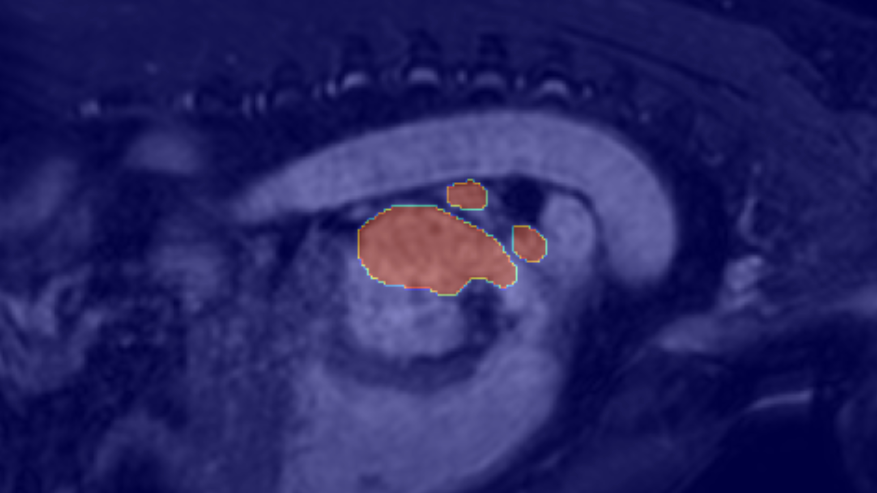}}\\
 &\multicolumn{4}{c}{Original Image} & \multicolumn{4}{c}{Label}\\
\toprule
& \textbf{UN} & \textbf{EUN} & \textbf{UN++} & \textbf{EUN++} & 
\parbox{1.2cm}{\centering \textbf{AHF\\UN}} & 
\parbox{1.2cm}{\centering \textbf{AHF\\EUN}} & 
\parbox{1.2cm}{\centering \textbf{UNT}} & 
\parbox{1.2cm}{\centering \textbf{EUNT}}\\
\midrule
\rotatebox{90}{\parbox{1.2cm}{\textbf{Pred}}} &
\includegraphics[width=0.115\textwidth]{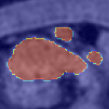} &
\includegraphics[width=0.115\textwidth]{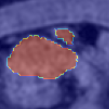} &
\includegraphics[width=0.115\textwidth]{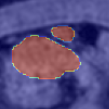} &
\includegraphics[width=0.115\textwidth]{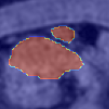} &
\includegraphics[width=0.115\textwidth]{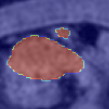} &
\includegraphics[width=0.115\textwidth]{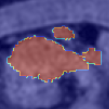} &
\includegraphics[width=0.115\textwidth]{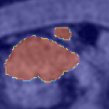} &
\includegraphics[width=0.115\textwidth]{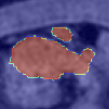}\\
\rule{0pt}{2pt}
\rotatebox{90}{\parbox{1.2cm}{\textbf{Conf}}}&
\includegraphics[width=0.115\textwidth]{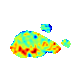} &
\includegraphics[width=0.115\textwidth]{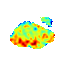} &
\includegraphics[width=0.115\textwidth]{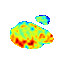} &
\includegraphics[width=0.115\textwidth]{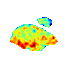} &
\includegraphics[width=0.115\textwidth]{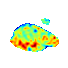} &
\includegraphics[width=0.115\textwidth]{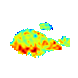} &
\includegraphics[width=0.115\textwidth]{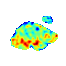} &
\includegraphics[width=0.115\textwidth]{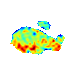}\\
\rotatebox{90}{\parbox{1.2cm}{\textbf{Exp.}}}&
\includegraphics[width=0.115\textwidth]{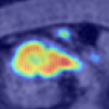} &
\includegraphics[width=0.115\textwidth]{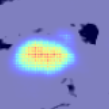} &
\includegraphics[width=0.115\textwidth]{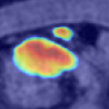} &
\includegraphics[width=0.115\textwidth]{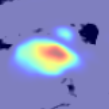} &
\includegraphics[width=0.115\textwidth]{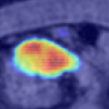} &
\includegraphics[width=0.115\textwidth]{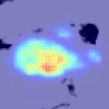} &
\includegraphics[width=0.115\textwidth]{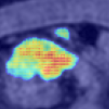} &
\includegraphics[width=0.115\textwidth]{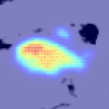}\\
\rotatebox{90}{\parbox{1.2cm}{\textbf{Uncert}}}&
 &
\includegraphics[width=0.115\textwidth]{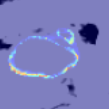} &
 &
\includegraphics[width=0.115\textwidth]{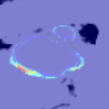} &
 &
\includegraphics[width=0.115\textwidth]{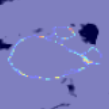} &
 &
\includegraphics[width=0.115\textwidth]{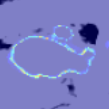}\\
\bottomrule
\end{tabular}
\caption{Comparison of different models on the MSD-Heart dataset. From top to bottom, the rows represent \textbf{predictions}, \textbf{confidence maps} (pixel-wise probabilities), \textbf{explanation maps} (Grad-CAM for U-Nets, MHEX+ for EU-Nets), and \textbf{uncertainty maps} generated by MHEX+'s gradient.}
\label{tab:one_sample}
\end{minipage}
\end{figure}

\subsection{Results}
In this section, we analyze model performance across datasets in Section~\ref{sec:Datasets}.

To ensure a fair comparison and reduce potential biases, we maintain consistent hyperparameters for all models, including image size, batch size, learning rate, and number of epochs. Each model is trained for up to 50 epochs with early stopping (patience = 10) and adaptive learning rate reduction (patience = 5). Consequently, training typically converges before the 30th epoch, with most trials stopping early. We perform five-fold cross-validation for all datasets.

EU-Nets consistently outperform their baseline counterparts, particularly in terms of stability, demonstrating robustness across different cross-validation folds and diverse data distributions. Moreover, integrating MHEX+ introduces only a minimal increase in model parameters (<0.1M). For overall performance comparison, see Table~\ref{tab:comparison}, and for individual sample results, refer to the \textit{Pred} row in Figure~\ref{tab:one_sample}.

\subsection{Salience Map}
Although both MHEX+ and Grad-CAM highlight important regions, they rely on fundamentally different mechanisms. \textbf{Grad-CAM} computes pixel-wise gradients, reproducing the segmentation output rather than providing deeper insights.

\begin{figure}[htbp]
\centering
    \begin{minipage}[t]{0.45\textwidth} 
        \vspace{0pt}
        \includegraphics[width=\linewidth]{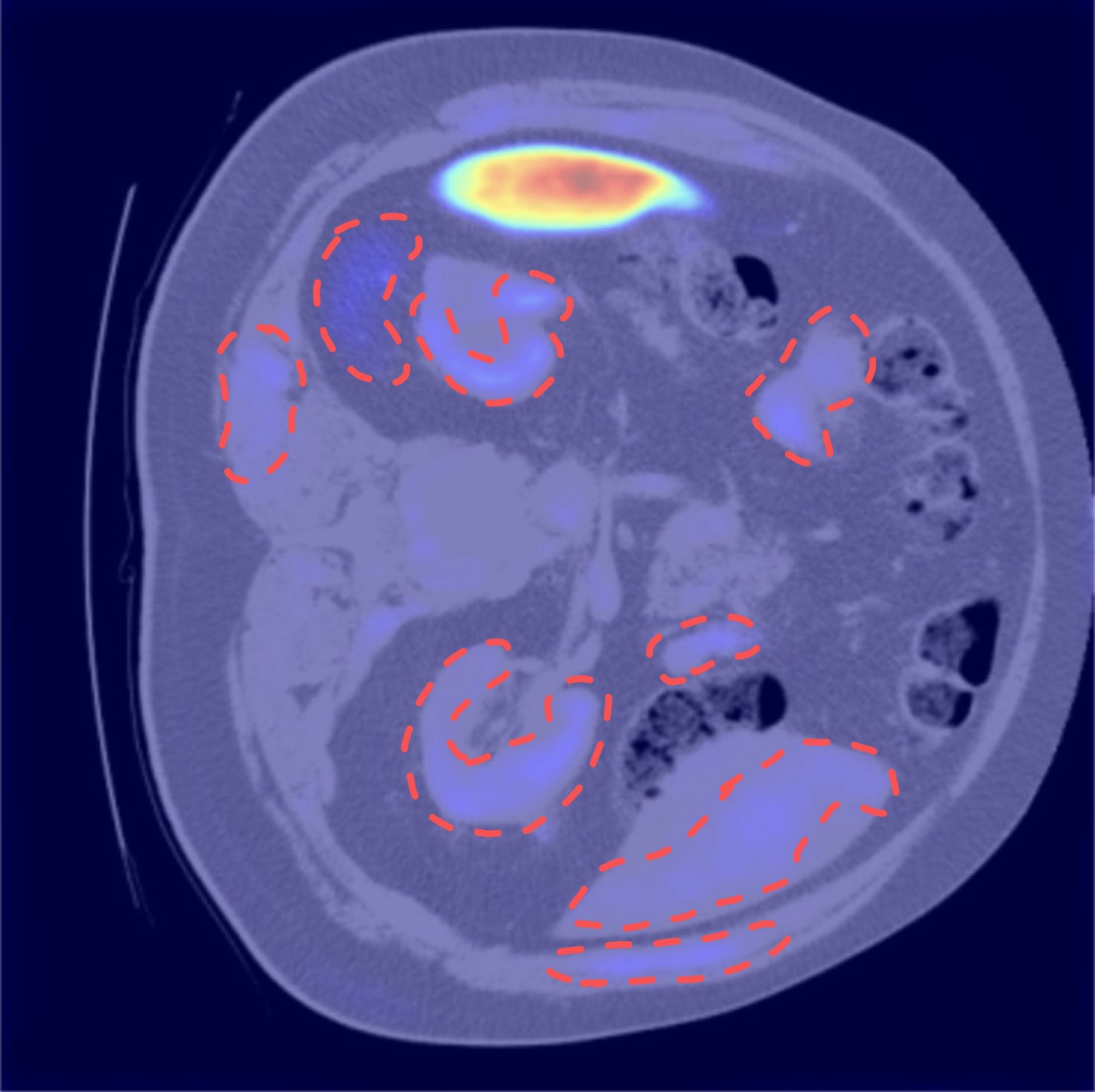}
        \label{fig:image1}
    \end{minipage}
\hspace{2mm}
    \begin{minipage}[t]{0.4\textwidth} 
    \centering

        \begin{minipage}[t]{0.55\linewidth}
            \vspace{1pt} \raggedright
            \includegraphics[width=\linewidth]{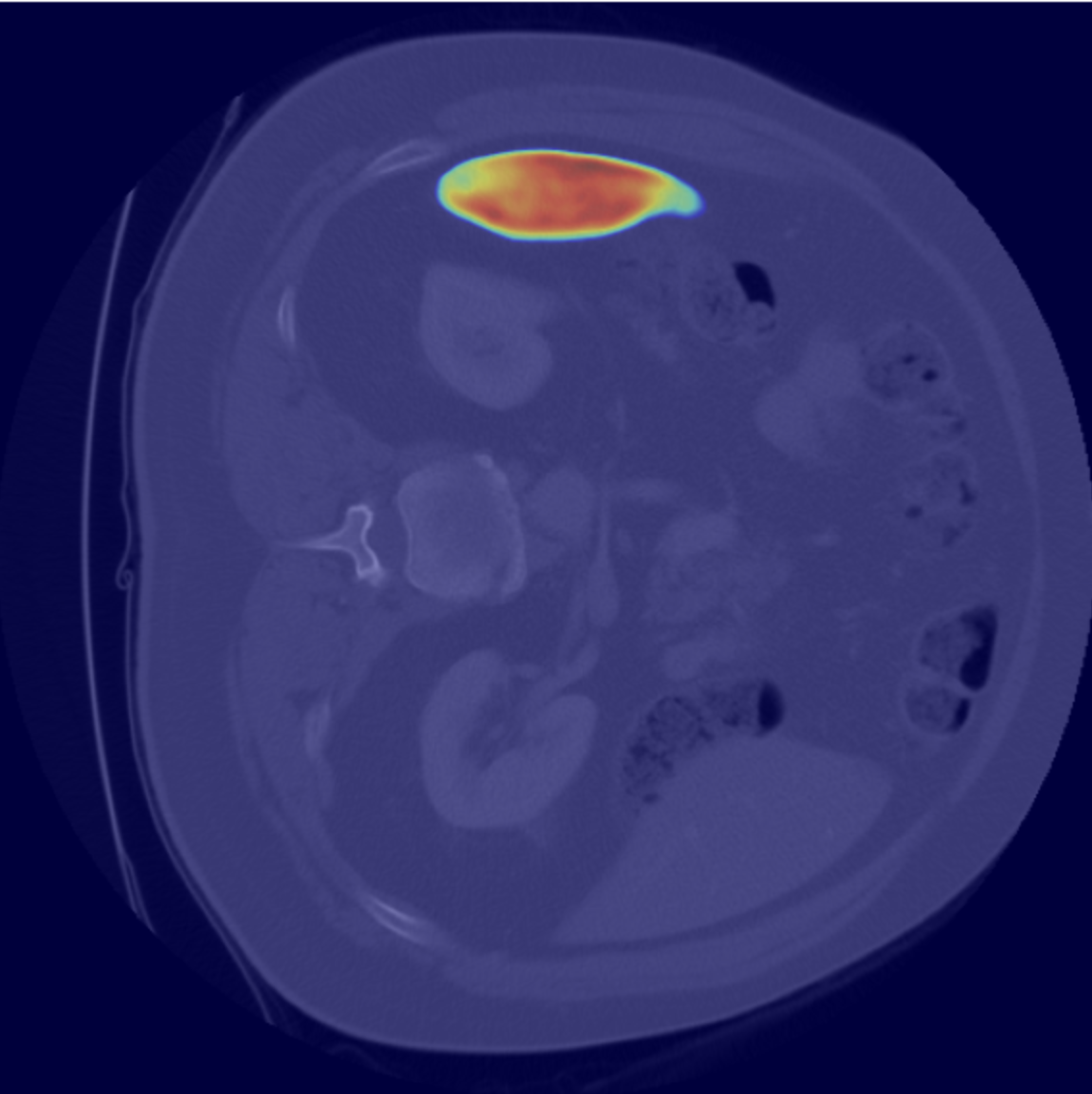}
            \label{fig:image2}
        \end{minipage}
        \begin{minipage}[t]{\linewidth}
            \vspace{-6pt} \raggedright
            \caption{Right: Grad-CAM salience map. Left: MHEX+ salience map, where deeper decisions, lightly activated, reveal decision traces and provide deeper insights. These regions are marked by dashed lines.}
            \label{fig:explaination_comparison}
        \end{minipage}
    \end{minipage}
\end{figure}

In contrast, MHEX+ leverages class-level equivalent convolution kernels to analyze how the network attends to each class. By aggregating learned weights instead of focusing solely on pixel-level gradients, MHEX+ captures multi-scale feature responses across decoder stages, highlighting relevant regions beyond the segmented target, such as liver and kidney areas. For comparison, see Figure~\ref{fig:explaination_comparison}. 

\subsection{Uncertainty Evaluation}
\label{sec:uncertainty_exp}

To evaluate the validity of our MHEX+ based uncertainty estimation (MU), we compare MU maps with Deep Ensemble uncertainty (DEU) maps (Section~\ref{sec:CollabGrad}).

\begin{figure}
    \centering
    \setlength{\tabcolsep}{5pt}
    \begin{tabular}{ccccc}
    \includegraphics[width=0.15\textwidth]{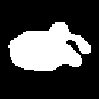} & 
    \includegraphics[width=0.15\textwidth]{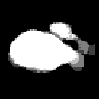}& 
    \includegraphics[width=0.15\textwidth]{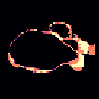}& 
    \includegraphics[width=0.15\textwidth]{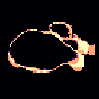} &
    \includegraphics[width=0.15\textwidth]{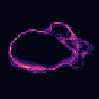}
    \\
         Label&  DE-Mean  & DE-Variance & DE-Entropy & MHEX-Uncert \\
    \end{tabular}
    \caption{Comparison of different uncertainty estimation methods. }
    \label{tab:uncertainty_visualization}
\end{figure}

We conducted experiments on the MSD-Heart dataset using 50 randomly selected samples and tested four proposed EU-Nets. Each model's uncertainty was computed via the MU approach, and the resulting maps were aggregated into a single composite visualization (Fig.~\ref{tab:uncertainty_visualization}, last column). For single uncertainty samples, see Figure~\ref{tab:one_sample}, last row. DEU maps (Fig.~\ref{tab:uncertainty_visualization}) were obtained using entropy- and variance-based uncertainty estimation (Section~\ref{sec:CollabGrad}).

To quantify the agreement between MU and DEU, we computed the IoU, Dice, and Pearson correlation between the uncertainty maps. The results, summarized in Table \ref{tab:uncertainty_comparison}, show statistically significant differences between MU and DEU (p value $= 0.0$), suggesting that MU captures uncertainty through internal gradient alignment within a single model.

\begin{table}[ht!]
    \setlength{\tabcolsep}{5pt}
    \centering
    \begin{tabular}{lccc}
        \hline
        \textbf{Method} & \textbf{IoU} & \textbf{Dice} & \textbf{Pearson Corr (p-value)} \\
        \hline
        \textbf{Entropy} & $0.3921 \pm 0.0619$ & $0.5607 \pm 0.0652$ & $0.6175 \pm 0.0352$ ($p=0.0000$) \\
        \textbf{Variance} & $0.3226 \pm 0.0575$ & $0.4852 \pm 0.0676$ & $0.5651 \pm 0.0344$ ($p=0.0000$) \\
        \hline\\
    \end{tabular}
    \caption{Comparison of Uncertainty Measures between MHEX+ and Deep Ensemble}
    \label{tab:uncertainty_comparison}
\end{table}

\section{Conclusion}
We propose EU-Nets, a family of explainable U-Net variants that enhance interpretability, segmentation performance, and uncertainty estimation. The MHEX+ module in EU-Nets is compatible with any U-Net-based architecture. Experiments on four medical imaging datasets demonstrate that EU-Nets consistently outperform their baselines. Compared to Grad-CAM, MHEX+ based saliency maps capture multi-scale and cross-decoder features, providing deeper insight into model predictions. Additionally, our gradient-based uncertainty estimation aligns well with Deep Ensemble methods, while requiring only a single model.

\bibliographystyle{splncs04}

\bibliography{ref}

\begin{thebibliography}{10}
\providecommand{\url}[1]{\texttt{#1}}
\providecommand{\urlprefix}{URL }
\providecommand{\doi}[1]{https://doi.org/#1}

\bibitem{marinov2024deepinteractivesegmentationmedical}
Marinov, Z., Jäger, P.F., Egger, J., Kleesiek, J., Stiefelhagen, R.: Deep interactive segmentation of medical images: A systematic review and taxonomy (2024), \url{https://arxiv.org/abs/2311.13964}

\bibitem{XAIinmedical}
Gipi{\v{s}}kis, R., Tsai, C.W., Kurasova, O.: Explainable ai (xai) in image segmentation in medicine, industry, and beyond: A survey. ICT Express  (2024)

\bibitem{wang2020score}
Wang, H., Wang, Z., Du, M., Yang, F., Zhang, Z., Ding, S., Mardziel, P., Hu, X.: Score-cam: Score-weighted visual explanations for convolutional neural networks. In: Proceedings of the IEEE/CVF conference on computer vision and pattern recognition workshops. pp. 24--25 (2020)

\bibitem{chattopadhay2018grad}
Chattopadhay, A., Sarkar, A., Howlader, P., Balasubramanian, V.N.: Grad-cam++: Generalized gradient-based visual explanations for deep convolutional networks. In: 2018 IEEE winter conference on applications of computer vision (WACV). pp. 839--847. IEEE (2018)

\bibitem{selvaraju2017grad}
Selvaraju, R.R., Cogswell, M., Das, A., Vedantam, R., Parikh, D., Batra, D.: Grad-cam: Visual explanations from deep networks via gradient-based localization. In: Proceedings of the IEEE international conference on computer vision. pp. 618--626 (2017)

\bibitem{gradisnotexplainable}
Suara, S., Jha, A., Sinha, P., Sekh, A.A.: Is grad-cam explainable in medical images? In: International Conference on Computer Vision and Image Processing. pp. 124--135. Springer (2023)

\bibitem{gradisnotexplainable2}
Teng, Z., Li, L., Xin, Z., Xiang, D., Huang, J., Zhou, H., Shi, F., Zhu, W., Cai, J., Peng, T., et~al.: A literature review of artificial intelligence (ai) for medical image segmentation: from ai and explainable ai to trustworthy ai. Quantitative Imaging in Medicine and Surgery  \textbf{14}(12), ~9620 (2024)

\bibitem{dodont}
Hasany, S.N., M{\'e}riaudeau, F., Petitjean, C.: The do’s and don’ts of grad-cam in image segmentation as demonstrated on the synapse multi-organ ct dataset. In: Medical Imaging with Deep Learning (2024)

\bibitem{subjectiveLogic}
J{\o}sang, A.: Subjective logic, vol.~3. Springer (2016)

\bibitem{DempsterShafer}
Dempster, A.P.: Upper and lower probabilities induced by a multivalued mapping. In: Classic works of the Dempster-Shafer theory of belief functions, pp. 57--72. Springer (2008)

\bibitem{gal2016dropout}
Gal, Y., Ghahramani, Z.: Dropout as a bayesian approximation: Representing model uncertainty in deep learning. In: international conference on machine learning. pp. 1050--1059. PMLR (2016)

\bibitem{lakshminarayanan2017simplescalablepredictiveuncertainty}
Lakshminarayanan, B., Pritzel, A., Blundell, C.: Simple and scalable predictive uncertainty estimation using deep ensembles (2017), \url{https://arxiv.org/abs/1612.01474}

\bibitem{mhex}
Sun, B., Liò, P.: Multi-head explainer: A general framework to improve explainability in cnns and transformers (2025), \url{https://arxiv.org/abs/2501.01311}

\bibitem{unet}
Ronneberger, O., Fischer, P., Brox, T.: U-net: Convolutional networks for biomedical image segmentation. In: Medical image computing and computer-assisted intervention--MICCAI 2015: 18th international conference, Munich, Germany, October 5-9, 2015, proceedings, part III 18. pp. 234--241. Springer (2015)

\bibitem{unet++}
Zhou, Z., Siddiquee, M.M.R., Tajbakhsh, N., Liang, J.: Unet++: A nested u-net architecture for medical image segmentation (2018), \url{https://arxiv.org/abs/1807.10165}

\bibitem{AHF-U-Net}
Munia, A.A., Abdar, M., Hasan, M., Jalali, M.S., Banerjee, B., Khosravi, A., Hossain, I., Fu, H., Frangi, A.F.: Attention-guided hierarchical fusion u-net for uncertainty-driven medical image segmentation. Information Fusion  \textbf{115},  102719 (2025). \doi{https://doi.org/10.1016/j.inffus.2024.102719}, \url{https://www.sciencedirect.com/science/article/pii/S1566253524004974}

\bibitem{hu2018squeeze}
Hu, J., Shen, L., Sun, G.: Squeeze-and-excitation networks. In: Proceedings of the IEEE conference on computer vision and pattern recognition. pp. 7132--7141 (2018)

\bibitem{woo2018cbam}
Woo, S., Park, J., Lee, J.Y., Kweon, I.S.: Cbam: Convolutional block attention module. In: Proceedings of the European conference on computer vision (ECCV). pp. 3--19 (2018)

\bibitem{U-NetTransformer}
Petit, O., Thome, N., Rambour, C., Soler, L.: U-net transformer: Self and cross attention for medical image segmentation (2021), \url{https://arxiv.org/abs/2103.06104}

\bibitem{lundberg2017unified}
Lundberg, S.: A unified approach to interpreting model predictions. arXiv preprint arXiv:1705.07874  (2017)

\bibitem{ribeiro2016whyitrustyou}
Ribeiro, M.T., Singh, S., Guestrin, C.: "why should i trust you?": Explaining the predictions of any classifier (2016), \url{https://arxiv.org/abs/1602.04938}

\bibitem{TTA}
Shanmugam, D., Blalock, D., Balakrishnan, G., Guttag, J.: Better aggregation in test-time augmentation (2021), \url{https://arxiv.org/abs/2011.11156}

\bibitem{lee2014deeplysupervisednets}
Lee, C.Y., Xie, S., Gallagher, P., Zhang, Z., Tu, Z.: Deeply-supervised nets (2014), \url{https://arxiv.org/abs/1409.5185}

\bibitem{li2022comprehensivereviewdeepsupervision}
Li, R., Wang, X., Huang, G., Yang, W., Zhang, K., Gu, X., Tran, S.N., Garg, S., Alty, J., Bai, Q.: A comprehensive review on deep supervision: Theories and applications (2022), \url{https://arxiv.org/abs/2207.02376}

\bibitem{schlemper2019attention}
Schlemper, J., Oktay, O., Schaap, M., Heinrich, M., Kainz, B., Glocker, B., Rueckert, D.: Attention gated networks: Learning to leverage salient regions in medical images. Medical image analysis  \textbf{53},  197--207 (2019)

\bibitem{he2015deepresiduallearningimage}
He, K., Zhang, X., Ren, S., Sun, J.: Deep residual learning for image recognition (2015), \url{https://arxiv.org/abs/1512.03385}

\bibitem{antonelli2022medical}
Antonelli, M., Reinke, A., Bakas, S., Farahani, K., Kopp-Schneider, A., Landman, B.A., Litjens, G., Menze, B., Ronneberger, O., Summers, R.M., et~al.: The medical segmentation decathlon. Nature communications  \textbf{13}(1), ~4128 (2022)

\bibitem{al2020dataset}
Al-Dhabyani, W., Gomaa, M., Khaled, H., Fahmy, A.: Dataset of breast ultrasound images. Data in brief  \textbf{28},  104863 (2020)

\bibitem{codella2018skin}
Codella, N.C., Gutman, D., Celebi, M.E., Helba, B., Marchetti, M.A., Dusza, S.W., Kalloo, A., Liopyris, K., Mishra, N., Kittler, H., et~al.: Skin lesion analysis toward melanoma detection: A challenge at the 2017 international symposium on biomedical imaging (isbi), hosted by the international skin imaging collaboration (isic). In: 2018 IEEE 15th international symposium on biomedical imaging (ISBI 2018). pp. 168--172. IEEE (2018)

\end{thebibliography}

\end{document}